\documentclass[lettersize,journal]{IEEEtran}
\usepackage{amsthm}
\newtheorem{theorem}{Theorem}

\newtheorem{definition}{Definition}
\newtheorem{remark}{Remark}
\newtheorem{assump}{Assumption}

\usepackage{amsmath,amsfonts,mathrsfs}
\usepackage{amsmath}               

\usepackage{algorithmic}
\usepackage{algorithm}
\usepackage{array}
\usepackage[caption=false,font=normalsize,labelfont=sf,textfont=sf]{subfig}
\usepackage{textcomp}
\usepackage{stfloats}
\usepackage{url}
\usepackage{verbatim}
\usepackage{graphicx}
\usepackage{cite}
\usepackage[utf8]{inputenc}
\usepackage{graphicx}
\usepackage{epstopdf}
\usepackage{etoolbox}
\usepackage{cite}
\usepackage{picinpar}
\usepackage{flushend}
\usepackage{bm}
\usepackage{xcolor}
\usepackage{indentfirst}
\usepackage[autostyle]{csquotes}

\begin{document}

\title{Underactuated Control of Multiple Soft Pneumatic Actuators via Stable Inversion}

\author{Wu-Te Yang$^{1}$,~\IEEEmembership{IEEE Student Member}, Burak K\"{u}rk\c{c}\"{u}$^{1}$,~\IEEEmembership{IEEE Member},\\ Masayoshi Tomizuka$^{1}$,~\IEEEmembership{IEEE Life Fellow}
\thanks{$^{1}$The authors are with the Department of Mechanical Engineering,
        University of California, Berkeley, MSC Lab, USA
        {\tt\small wtyang; bkurkcu; tomizuka@berkeley.edu}}}



\maketitle

\begin{abstract}
Soft grippers, with their inherent compliance and adaptability, show advantages for delicate and versatile manipulation tasks in robotics. This paper presents a novel approach to underactuated control of multiple soft actuators, explicitly focusing on the coordination of soft fingers within a soft gripper. Utilizing a single syringe pump as the actuation mechanism, we address the challenge of coordinating multiple degrees of freedom of a compliant system. The theoretical framework applies concepts from stable inversion theory, adapting them to the unique dynamics of the underactuated soft gripper. Through meticulous mechatronic system design and controller synthesis, we demonstrate the efficacy and applicability of our approach in achieving precise and coordinated manipulation tasks in simulation and experimentation. Our findings not only contribute to the advancement of soft robot control but also offer practical insights into the design and control of underactuated systems for real-world applications.
\end{abstract}


\section{Introduction}
\IEEEPARstart{S}{oft} robotics has become an emerging field, offering solutions for tasks that traditional robots struggle to accomplish. In particular, the inherent compliance and adaptability of soft actuators show advantages for applications requiring delicate manipulation~\cite{navas2021gripper, george2020survey} and interaction with complex or unknown environments~\cite{fumiya2011review, yang2022sensor}. Compared to rigid-bodied robotic hands, soft grippers stand out for their ability to conform to a wide range of object shapes and sizes, making them indispensable in domains such as medical robots~\cite{alici2018bending} and human-robot interactions~\cite{demir2020design}. However, achieving precise control over the motion and coordination of multiple soft fingers within a gripper remains a challenge~\cite{yang2023control}. The soft fingers in this paper are pneumatically driven, and the air is supplied by air pumps. If the number of fingers exceeds the number of pumps, the system is under-actuated. Since the air pump is bulky, it is desired to minimize the number of air pumps~\cite{underactuated2019he,tuna2019syn,pump2024zhou}.


\begin{figure}[t]
    \centering
    \includegraphics[width=210pt]{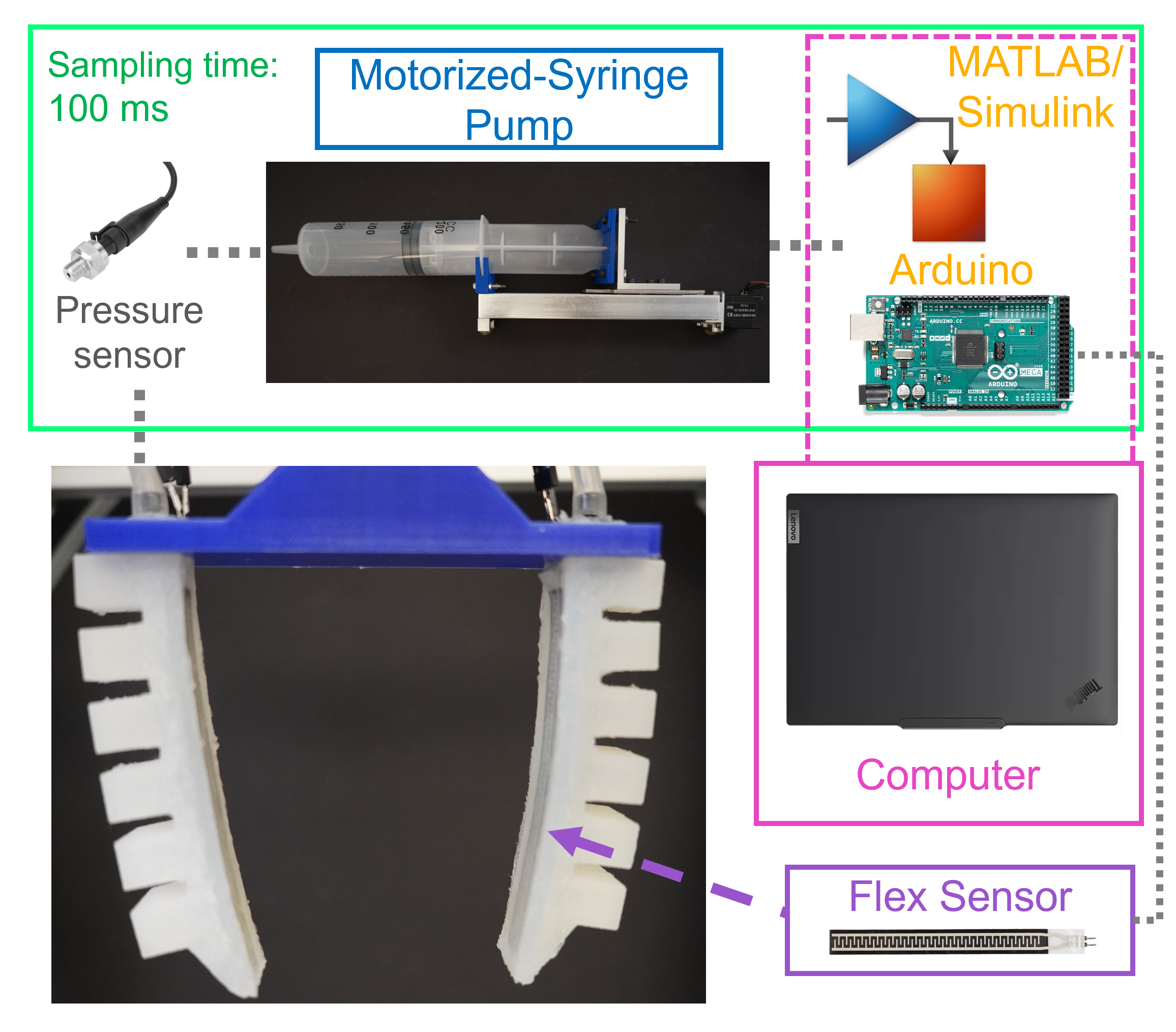}
    \caption{The soft gripper has two fingers and is driven by a single syringe pump to achieve underactuated control via model inversion. The control commands are generated in MATLAB\textregistered/Simulink and are converted to PWM for the stepper motor in the syringe pump. The bending angles of both fingers are measured by the flex sensor embedded in each soft gripper.}
    \label{fig_1}
\end{figure}

Despite the recent development of soft robot control, achieving coordinated control under the underactuated control framework for soft robots is seldom discussed and remains a challenge. A couple of works addressed the soft robot control issues by applying nonlinear controllers~\cite{skorina2015nonlinear, skorina2016nonlinear}, adaptive controllers~\cite{skorina2017adaptive, tang2021adaptive}, and optimal controllers~\cite{yang2023control, best2016mpc}. Those control strategies enable high-performance control of soft robots with high degrees of freedom. However, the control of these systems becomes increasingly complex as the number of degrees of outputs exceeds that of inputs (underactuated control)~\cite{underactuated2019he}. On the other hand, soft materials have an uncertain nature, and thus soft robots exhibit model uncertainty~\cite{rebecca2015material}. 
Various sources contribute to these uncertainties in modeling soft robots, which can be categorized into two types epistemic and statistical uncertainties~\cite{kim2021prob}.
Uncertainty can also arise from changes in the physical properties of soft materials.  For example, polymer materials are susceptible to aging. Actuators or sensors made of polymer need time to fully cure and achieve chemical stability, during which their properties gradually shift~\cite{jung2020spa}.
Furthermore, prolonged use can alter the internal bonding structure of these materials~\cite{gao2019actuator}.
Variations in the materials themselves and in manufacturing tolerances add another layer of uncertainty. When soft robots operate in unstructured environments~\cite{pin2022soro}, the robustness of the controller becomes more critical. Addressing this challenge may require the development of novel control strategies that can effectively coordinate multiple soft actuators within a soft gripper, enabling precise and adaptive manipulation tasks in real-world scenarios.

The primary objective of this paper is to develop control algorithms for achieving coordination in multi-finger soft grippers, which are modeled as single-input-multi-output (SIMO) systems.
Our approach integrates both feedforward and feedback control loops. The feedforward control mechanism incorporates a stable model inversion technique that effectively controls the motions of multiple soft fingers. Given the inherent uncertainty of soft materials, the feedback loop is adept at mitigating unexpected errors, noise, or disturbances that may arise between the mathematical model and the real system. Comprehensive validation of the control algorithms is conducted through simulations and experimentation. The theoretical framework underpinning these control algorithms is initially introduced in~\cite{burak2024algebraic}, where its efficacy is established by necessary and sufficient conditions. The contributions of this research lie in the innovative application of stable inversion control algorithms to address uncertain soft robotic systems. Notably, the proposed controller achieves coordination of multi-finger soft grippers with a single input, thereby demonstrating the applicability of these control algorithms to SIMO control problems.

To position our contributions, we compare our research with recent works. In our prior study~\cite{yang2023control}, we employed individual syringe pumps for each finger within a multi-finger soft gripper (full-drive) to attain precise and synchronized motions, so the number of air pumps is equal to the number of fingers. In contrast, the present research adopts stable model inversion alongside a single air pump to achieve coordination across all fingers within the multi-finger soft gripper. 
The algebraic-related method was proposed in~\cite{keppler2022under}, which established input coordination transformation that made the underactuated soft robotic systems become quasi-fully actuated systems. Although our feedforward control uses a similar concept, there is a feedback loop to cope with the soft robotic systems' uncertainty that ensures the system's robustness. Pustina et al.~\cite{pustina2022under} studied the controllability and stability of the underactuated soft robots. But our work focuses on the SIMO problem of coordinating motions of multiple soft robots with a single input. Another major contribution is the experimental investigation and revelation of the parameter-varying nature of uncertainties in soft actuators. Furthermore, after conducting several experiments, a correlation between the uncertainty envelope and the bandwidth of underactuated control is established. Overall, this research studies the stable underactuation of soft robots with robust performance. 

The remainder of this paper is organized as follows. Section~\ref{sec2} introduces the mathematics preliminary and problem statement. Section~\ref{sec3} describes the full mechatronic design and proposed controller design. Section~\ref{sec4} evaluates the feasibility and applicability of the controller by simulations and experimentation. Section~\ref{sec5} discusses the experimental results and concludes the work.

\section{Problem Formulation}\label{sec2}
\subsection{Mathematics Preliminaries}\label{subsec21}

The set $\Bbb{R}$ is a real number field. 
Then, the set of all these rational functions in $s$ over $\Bbb{R}$ forms a field, denoted by $\Bbb{R}(s)$~\cite{forney1975vs}.
The sets of $n_y \times n_u$ matrices with elements in $\Bbb{R}$ and $\Bbb{R}(s)$ are denoted by $\Bbb{R}^{n_y \times n_u}$ and $\Bbb{R}(s)^{n_y \times n_u}$ respectively. 

The rank of a matrix $P$ over the field $\Bbb{R}(s)$ is defined as the maximum number of linearly independent subsets of its columns (or rows) \cite{anto2005minimal}. This is denoted by $\textnormal{rank}_{\Bbb{R}(s)}(P)$. A set of vectors $v_1, v_2, ..., v_{n_u}$ is linearly independent in the field $\Bbb{R}(s)$ if and only if the condition $\sum a_i v_i = 0$ and $a_i = 0$ for every $i$ with the scalars $a_i$ in $\Bbb{R}(s)$.

Suppose the rank of $P(s) \in \Bbb{R}(s)^{n_y \times n_u}$ is $r$ where $1\leq r \leq \text{min}(n_u,n_y)$. By choosing $P(s)$'s $r$ linearly independent columns, we can form a full-rank rational matrix as
\begin{equation}
\begin{split}
\mathcal{L}(P) = \{p_i \ | \ 1\leq i \leq r\}
\label{eqn_LD}
\end{split}
\end{equation}
with $p_i$ being linearly independent over $\Bbb{R}(s)$. Now, we can define the \textit{Image (Range) Set} for the real-rational matrices in $s$ as follows. 
\begin{equation}
\begin{split}
\textnormal{Im}_{\Bbb{R}(s)}(P) = \biggl\{\sum_{i=1}^{r} {c_i}{p_i}: c_i\in \Bbb{R}(s), p_i \in \mathcal{L}(P) \biggl\}\subseteq \Bbb{R}(s)^{n_y \times 1}
\label{eqn_image}
\end{split}
\end{equation}

\begin{theorem} (Rouche-Capelli Theorem) \label{thm_RC}
Consider $P \in \Bbb{R}(s)^{n_y \times n_u}$ with $\textnormal{rank}_{\Bbb{R}(s)}(P)=r$ and $Y \in \Bbb{R}(s)^{n_y \times 1}$. The solution(s) $U$ for the equation $PU=Y$ is exist if and only if
\begin{align}
    \textnormal{rank}_{\Bbb{R}(s)}(P) = \textnormal{rank}_{\Bbb{R}(s)}\underbrace{([P(s)~:~Y(s)])}_{\in \Bbb{R}(s)^{n_y \times (n_u+1)}}=r
\end{align}
\end{theorem}

Some notations used in the paper are: the $\Re(\cdot)$ represents the real part of the given complex number, and $\Im(\cdot)$ denotes the imaginary part of the number. $\mathcal{L}_{\infty}(j\Bbb{R})$ represents functions bounded on $\Re(s)$ = 0 including at $\infty$, and $\mathcal{RH}_{\infty}$ is the Hardy Space and denotes the set of $\mathcal{L}_{\infty}(j\Bbb{R})$ functions analytic in $\Re(s) > 0$.


\subsection{Problem Statement}\label{subsec22}
The soft gripper, equipped with multiple fingers that can provide a human-like grasping experience, serves as the plant for the proposed control algorithms. Soft robotic systems, by their nature, exhibit nonlinear behaviors due to the compliance and deformability of their materials \cite{rebecca2015material}. However, the nonlinearity of certain soft materials may not be evident under reasonably constrained deformations. Therefore, within the constraints of deformations, it is feasible to approximate the behavior of soft robots using linear uncertainty models \cite{yang2023control, yang2024design}.

The multi-fingered soft gripper can be considered a type of multi-input-multi-output (MIMO) linear time-invariant (LTI) system where the state-space realization follows as

\begin{equation}
\begin{split}
&{\dot x} =  A {x} + B {u} \\
&{y} = C {x}, x(0) = 0
\label{eqn_mimo}
\end{split}
\end{equation}
where $A \in \Bbb{R}^{n\times n}$, $B \in \Bbb{R}^{n\times n_u}$, $C \in \Bbb{R}^{n_y\times n}$, $x(t)$ $\in \Bbb{R}^{n\times 1}$, $y(t) \in \Bbb{R}^{n_y \times 1}$, and $u(t) \in \Bbb{R}^{n_u \times 1}$. 

Equivalently, by using the Laplace transformation, (\ref{eqn_mimo}) can be expressed with Transfer Function Matrices (TFM) as follows

\begin{equation}
\begin{split}
    Y(s) & =  C(s)(sI - A)^{-1}BU(s) \\
    \Rightarrow Y(s) &= P(s)U(s)
\label{eqn_simo}
\end{split}
\end{equation}

\begin{assump}\label{ass_minimumsystem} The system given in (\ref{eqn_simo}) is assumed to have the following properties:
\begin{enumerate}
    \item[a.] The system (\ref{eqn_simo}) is minimal, implying controllability and observability.
    \item[b.] The system (\ref{eqn_simo}) is the minimum phase and Hurwitz, meaning all poles and transmission zeros are in the left half of the complex plane.
    \item[c.] $Y(s) \in \textnormal{Im}_{\Bbb{R}(s)}(P)$, indicating that any output function is in the range space of $P$, making the output function achievable.
\end{enumerate}
\end{assump}
These assumptions ensure that the system is well-posed and stable, facilitating the design and analysis of the control strategies. These assumptions will be validated through analysis and assess their applicability under real-world conditions in Sec.~\ref{pre_eval}.

To achieve a less complicated design, enhanced energy efficiency, and reduced weight, we can use a single input source to control all fingers with different dynamics and outputs. With this approach (having one input), the system described in (\ref{eqn_simo}) becomes a SIMO system

\begin{align}
\begin{bmatrix} Y_{1}(s)\\ \vdots \\ Y_{n_y}(s) \end{bmatrix} = \begin{bmatrix} P_1(s)\\ \vdots \\ P_{n_y}(s) \end{bmatrix} {U(s)} 
\label{eqn_simo2}
\end{align}
where $P_i(s) \in \Bbb{R}(s)$, $Y_i(s) \in \Bbb{R}(s)$, $i = 1 ... n_y$, and $U(s) \in \Bbb{R}(s)$. 
Solving the algebraic equality for $U(s)$ gives us the \textit{exact left inverse}. Since $P(s)$ (where $P(s) \ne 0)$ consists of a single-column real rational vector, we always have $\textnormal{rank}_{\Bbb{R}(s)}(P)=1$. Together with \textit{Assumption \ref{ass_minimumsystem}. c.}, the condition for the existence stated in \textit{Theorem~\ref{thm_RC}} as
\begin{align}
    \textnormal{rank}_{\Bbb{R}(s)}(P) = \textnormal{rank}_{\Bbb{R}(s)}{([P(s)~:~Y(s)])}=1
\end{align}
is met, which ensures that (\ref{eqn_simo}) has a unique solution. On the other hand, since $P(s)$ has no unstable invariant zeros (\textit{Assumption~\ref{ass_minimumsystem} b.}), the unique solution is also stable, i.e., U(s) $\in \mathcal{RH}_\infty$.

\section{Methodology}\label{sec3}
In this section, we thoroughly introduce the mechatronic system design including the soft pneumatic actuators~\cite{yang2023design, yang2024design} and syringe pump design~\cite{yang2023pump}. The research delves into the study of a soft gripper system formed by integrating those components. The soft pneumatic actuators serve as the fingers driven by the syringe pump and the proposed control algorithms. The dynamical modeling of the systems based on mechanics and fluid dynamics is presented in Sec.~\ref{subsec32} and the controller is designed in Sec.~\ref{subsec33}.

\begin{figure*}
  \centering
  \includegraphics[width=\textwidth]{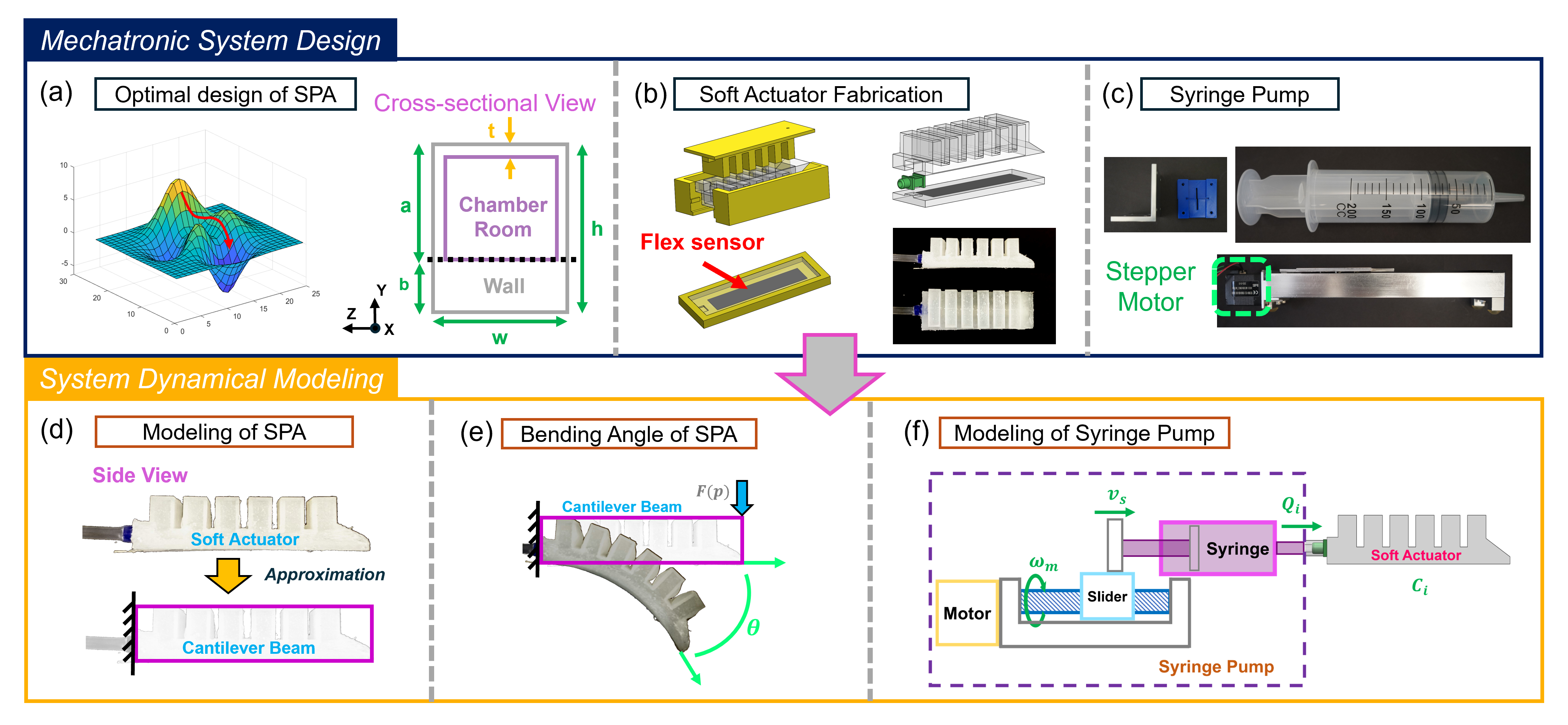}
  \caption{The design of the mechatronic system can be seen in (a), (b), and (c), while the system modeling is observed in (d), (e), and (f). (a) visualizes how the optimal dimensional parameters are searched in a non-convex space. (b) illustrates the fabrication process of the soft pneumatic actuator and the flex sensor is embedded during the fabrication process. (c) shows the appearance of the syringe pump, and it is made of a commercial linear actuator and a commercial syringe. The (d) and (e) visualize how the structure of the soft actuator is approximated by a cantilever beam and how the bending angle is measured. The modeling schematic of the syringe pump is displayed in (f).}
  \label{fig_2}
\end{figure*}

\subsection{Mechatronic Design}\label{subsec31}
The mechatronic system design is illustrated in Fig.~\ref{fig_2} (a), (b), and (c). The soft gripper is composed of two main components: soft fingers and a syringe pump, as in Fig.~\ref{fig_1}. The design methodology of each component will be elaborated in the following paragraphs.

\subsubsection{Soft Actuator Design}\label{subsubsec311}
The soft actuator is designed under an optimal model-based design framework that considers force/torque, bendability, and controllability simultaneously during the design stage~\cite{yang2024design}. The dimensional parameters of a soft pneumatic actuator are as shown in Fig.~\ref{fig_2} (a), the cross-sectional view of the soft actuator's chamber room. The optimal dimensional parameters are searched by the optimization framework below

\begin{equation}
    \begin{split}
    \max_{a,b,w,t} & {~\Bar{T}(p) + \Bar{\theta}(p)} \\
    \textrm{s.t.} &~\dot p = 0\\
    &{a_1} \leq a \leq {a_2}    \\
    &{b_1} \leq b \leq {b_2}    \\
    &{h_1} \leq a+b \leq {h_2}    \\
    &{C_1} \leq {Ew(a+b)^{n+2}} \leq {C_2}    \\
    \label{eqn_opt}
    \end{split}
\end{equation}
where $a$ is the top of the chamber to the neutral surface, $b$ is the neutral surface to the bottom of the chamber, $a+b$ is the height of the soft actuator as in Fig.~\ref{fig_2} (a), $E$ is Young's modulus of the selected material, and $n$ is a data-driven parameter related to soft materials~\cite{lee2002large}. Note that $w$ and $t$ represent the width and wall thickness of the cross-sectional area (Fig.~\ref{fig_2} (a)). However, the optimal values for these two parameters tend to reach their respective upper and lower bounds, so they are not included in the (\ref{eqn_opt}) and are determined by the designer.

The $\Bar{T}(p)$ represents the Pressure-to-Force/Torque model, which is obtained by mechanics analysis of the soft actuator~\cite{yang2024design}, while the $\Bar{\theta}(p)$ stands for the Pressure-to-Bending model which is derived by a nonlinear mechanics theory~\cite{lee2002large, yang2024design}. Both $\Bar{T}(p)$ and $\Bar{\theta}(p)$ are functions of the dimensional parameters, $a$, $b$, $w$, and $t$ as shown in Fig.~\ref{fig_2} (a). There exists an optimal parameter set that maximizes the objective function of~(\ref{eqn_opt}), namely $\Bar{T}(p) + \Bar{\theta}(p)$~\cite{yang2024design}. The parameter set is searched by optimization algorithms. The constraint of $Ew(a+b)^n$ aims to place the natural frequency of the soft actuator in the desired range. The remaining parameters that are not considered in the~(\ref{eqn_opt}) include the Young modulus, length of the structure, and number of chamber rooms. They will be discussed in the following paragraph.

The range of dimensional parameters is selected by referencing the size of human fingers~\cite{wang2019actuator}, so the constraint of $a$, $b$, and the value of $w$ are determined and $n$ is decided by the selected soft material. To position the natural frequency in the desired range (2 - 3 $rad/s$), the Smooth-on Ecoflex\textregistered ~Dragon Skin 20 is selected and its Young's modulus is 0.34 $MPa$. The length of the soft actuator and the number of chambers are coupled. The more the number of chambers, the longer the length. The length of 100 $mm$ is chosen to avoid the buckling effect caused by the long structure and the corresponding number of the chamber rooms is 6. 

The soft actuator is fabricated by two molds as illustrated in Fig.~\ref{fig_2} (b). There are upper and bottom components on the left side of Fig.~\ref{fig_2} (b). The Smooth-on Ecoflex\textregistered~Dragon Skin 20 is in the liquid state, and its curing time is around 4 hours. A flex sensor is embedded into the bottom component, as shown in Fig.~\ref{fig_2} (b), before the liquid rubber becomes a solid state. When the two components are removed from the molds, they are bonded by the silicone adhesive Smooth-on Sil-poxy\textregistered, as shown in the top right of Fig.~\ref{fig_2} (b). The appearance of the soft actuator is as shown in the bottom right of Fig.~\ref{fig_2} (b).

\subsubsection{Syringe Pump Design}
The schematic of the syringe pump is shown in Fig.~\ref{fig_2} (c), which is used to pressurize soft pneumatic actuators. The design of the syringe pump attempts to reduce the complexity of the pressure control and reduce the weight and size compared to traditional air pumps. The syringe pump, inspired by the hydraulic system, is made of a commercial syringe and a commercial linear actuator. The syringe pump is driven by the linear motor in the linear actuator~\cite{yang2023pump}. The pressure is adjusted by controlling the position of the slider. 

The precision of the linear actuator and the volume of the syringe have an influence on the accuracy and controllability of the syringe pump. The linear actuator, Fulride by NSK Ltd., and a syringe with 150 $mL$ are chosen to fabricate the syringe pump. The accuracy of the Fulride could be $\mu m$ scale and the volume of the syringe could provide pressurize up to three soft actuators to generate $\pi/2$ $rads$. Some 3D-printed components, which are easy to make and lightweight, are designed and manufactured to assemble the syringe and the linear actuator.

\subsubsection{Multi-finger Soft Gripper}
Multiple soft pneumatic actuators and the syringe pump form a soft gripper module, which is assembled by 3D-printed connectors and rubber tubes. The detailed compositions of the soft gripper, including sensor setup, will be described in Sec.~\ref{subsec42}.

\subsection{System Modeling}\label{subsec32}
Prior to designing the underactuated controller for the soft gripper, we need the full system dynamical model of both the soft actuators and the syringe pump. The models of soft actuators and the syringe pump are cascaded to obtain the full system model matrix. The model of each component is developed in the following subsection.

\subsubsection{Modeling Soft Actuators}
The dynamical model of the soft pneumatic actuator is obtained by modeling its approximated structure as shown in Fig.~\ref{fig_2} (d). Due to the large bending nature of the soft actuator, a nonlinear second-order model is utilized to model and capture its motions. The nonlinear second-order model is described as~\cite{yang2024model}

\begin{align}
    \begin{split}
    {M_{eq}} \ddot{\theta} + C_n{\dot \theta} + K_n {\theta}^{n+\Delta n} = F(p)
    \label{eqn_14}
    \end{split}
\end{align}
where $n$ is the data-driven parameter and is the same as the $n$ in (\ref{eqn_opt}), $\Delta n$ is caused by the uncertainty of soft materials, $\theta$ is the bending angle as shown in Fig.~\ref{fig_2} (e), $F(p)$ is the equivalent force generated by the input pressure $p$ from the syringe pump, $M_{eq}$ is the equivalent mass of the soft actuator, $C_n$ is the damping constant of the soft actuator. The $M_{eq}$ and $C_n$ currently are estimated by performing system identification applied to experimentally obtained responses in MATLAB\textregistered, which has an average accuracy of approximately $95.3 \%$. $K_n$ is the spring constant expressed as

\begin{align}
    \begin{split}
    K_n = (\frac{n+1}{n})^n(\frac{EI_n}{L_0^{n+1}})
    \label{eqn_11}
    \end{split}
\end{align}
where $L_0$ is the initial length of the soft actuator, $n$ is the same as the parameter in~(\ref{eqn_opt}) and it determined by either experiments or data-driven approach~\cite{yang2024model}, $I_n$ is the modified moment of inertia for a large deflection component~\cite{lee2002large} and it is expressed as

\begin{align}
    \begin{split}
    {I_n} = (\frac{1}{2})^n(\frac{1}{2+n})w({a+b})^{(2+n)} 
    \label{eqn_12}
    \end{split}
\end{align}
If $n = 1$ and $\Delta n = 0$, the dynamical equations~(\ref{eqn_14}), (\ref{eqn_11}), and (\ref{eqn_12}) become linear equations. Note that the linear model can capture the behavior of the real system within $\theta = 0$ to $4\pi/9$ $rad$~\cite{yang2024model}.
In this study, the actuator is aimed to remain within this specific range. Therefore, we set $n = 1$ in (\ref{eqn_14}), which yields:

\begin{align}
    \begin{split}
    \ddot{\theta} + (C_n/{M_{eq}}) {\dot \theta} + (K_n/{M_{eq}})  {\theta} = c\cdot p/{M_{eq}} 
    \label{eqn_16}
    \end{split}
\end{align}

The state-space form is, therefore, written as

\begin{align}
A_1 = \begin{bmatrix} 0 & 1 \\ -\frac{K_n}{M_{eq}} & -\frac{C_n}{M_{eq}} \end{bmatrix},
B_2 = \begin{bmatrix} 0 \\ 1\end{bmatrix},
C_2 =\begin{bmatrix} \frac{c \cdot p}{M_{eq}} \\ 0 \end{bmatrix}^T
\label{eqn_ss_1}
\end{align}
where $F(p)$ is represented as $c\cdot p$, $c$ is a constant affected by $a$, $b$, $w$, and $t$~\cite{yang2023design}. If we stack the system model of $n_y$ fingers, the state-space form becomes

\begin{align}
A_{stk} = \begin{bmatrix} A_1 & 0 & 0 \\ 0 & \ddots & 0 \\ 0 & 0 & A_{n_y} \end{bmatrix},
B_{stk} = \begin{bmatrix} B_1 \\ \vdots \\ B_{n_y}\end{bmatrix},
C_{stk} =\begin{bmatrix} {C_1}^T \\ \vdots \\{C_{n_y}}^T \end{bmatrix}
\label{eqn_ss_2}
\end{align}
According to (\ref{eqn_simo2}), we get the system equation in the Laplace domain. Thus, the system model matrix~$P(s)^{n_{y}\times 1}$ is described as

\begin{align}
{P}(s) = \begin{bmatrix} \frac{c\cdot p/M_{eq}}{s^2 + (C_{n\_1}/M_{eq}) s + K_{n\_1}/M_{eq}}\\ \vdots  \\ \frac{c\cdot p/M_{eq}}{s^2 + (C_{n\_n_{y}}/M_{eq}) s + K_{n\_n_{y}}/M_{eq}} \end{bmatrix}
\label{eqn_17}
\end{align}
where the equivalent mass of the soft actuators is almost the same and the same symbol $M_{eq}$ is used. Although the two fingers share the same dimensional parameters such as height, weight, etc., their $C_n$ and $K_n$ in (\ref{eqn_16}) are slightly different due to the uncertainty of soft materials. 

\begin{equation} 
\begin{split}
    C_1 &\leq C_n \leq C_2\\
    K_1 &\leq K_n \leq K_2
\end{split}
\label{eqn_uncertain}
\end{equation}
This nature leads to different motions when a feedback controller is applied. The asynchronized motions further lead to grasping failure\cite{yang2023control}. One of the aims of this study is to address such grasping failures by utilizing the algebraic stable inversion approach for SIMO setting, and the experimental results will be presented in Sec.~\ref{subsec44}.

\subsubsection{Modeling Syringe Pump}
The configuration of the syringe pump is visualized in Fig.~\ref{fig_2} (f). The dynamical modeling of the syringe pump starts at the linear motor. When the linear motor works, it will move the slider

\begin{align}
    \begin{split}
    v_s = \frac{l}{2\pi} \omega_m
    \label{eqn_18}
    \end{split}
\end{align}
where $v_s$ is the speed of the motor, $l$ is the lead of the screw inside the linear actuator, and $\omega_m$ is the motor speed. The $v_s$ times the inner cross-sectional area of the syringe becomes the output air flow rate

\begin{align}
    \begin{split}
    Q_i = A_s v_s = \frac{l}{2\pi} \omega_m
    \label{eqn_19}
    \end{split}
\end{align}
where $A_s$ is the inner cross-sectional area of the syringe, and $Q_i$ is the output air flow rate of the syringe. When $Q_i$ is divided by the capacity of the soft actuator, it becomes the pressure-changing rate inside the soft actuator

\begin{align}
    \begin{split}
    \dot p = \frac{Q_i}{C_i} = \frac{A_sl}{2\pi C_i} \omega_m
    \label{eqn_20}
    \end{split}
\end{align}
The dynamics of the syringe pump is the first-order system. The maximum angular velocity of the motor is 5 $rev/s$. The $C_i$ will expand as it is pressurized; however, its effect can be ignored as the input pressure is below 0.1 $MPa$ and the bending angle of the soft actuator is below $2\pi/3$ $rad$. The $C_i$ here is considered as a constant.

The full model of a single soft actuator is the cascade of the (\ref{eqn_16}) and (\ref{eqn_20}) 

\begin{align}
A_1 = \begin{bmatrix} 0 & 1 & 0\\0 & 0 & 1\\ 0 & -\frac{K_n}{M_{eq}} & -\frac{C_n}{M_{eq}} \end{bmatrix},
B_1 = \begin{bmatrix} 0 \\0 \\ 1\end{bmatrix},
C_1 =\begin{bmatrix} \frac{c\cdot pA_sl}{2\pi C_i M_{eq}} \\ 0 \\ 0\end{bmatrix}^T
\label{eqn_ss_full}
\end{align}
The resulting systems from $\omega_m$ to $\theta$ are third-order and have a pole at the imaginary axis. Hence, the full system model matrix $P(s)^{n_{y} \times 1}$ in Laplace domain can be obtained by referencing (\ref{eqn_simo2}) and (\ref{eqn_ss_2})

\begin{align}
{P(s)} = \begin{bmatrix} \frac{c\cdot pA_sl/2\pi C_i M_{eq}}{s^3 + (C_{n\_1}/M_{eq}) s^2 + (K_{n\_1}/M_{eq})s}\\ \vdots \\ \frac{c\cdot pA_sl/2\pi C_i M_{eq}}{s^3 + (C_{n\_n_{y}}/M_{eq}) s^2 + (K_{n\_n_{y}}/M_{eq})s} \end{bmatrix}
\label{eqn_22}
\end{align}
The full system matrix is causal, and the minimum phase and the $P(s)$ is full rank and invertible.

Even though all finger angles ($\theta$) remain within $[0, 4\pi/9]$, equation (\ref{eqn_ss_full}) cannot accurately capture the exact behavior of the soft gripper due to its inherent uncertainty structure~\cite{kim2021prob}. To achieve a more accurate representation, we can modify the model in (\ref{eqn_16}) using a multiplicative uncertainty approach as in~\cite{zhou1998robust}, defining a set of all possible plants for each finger as follows:

\begin{equation}
\Pi := \{(I+{\Delta} {W_T})P~|~\forall \parallel \Delta \parallel_{\infty}\leq \gamma\}
\label{eqn_pertub}
\end{equation}

Here, the transfer function $W_T \in \mathcal{RH}_{\infty}$ represents the spatial and frequency characteristics of the uncertainty. $\Delta$ denotes any unstructured and unknown yet stable function~\cite{zhou1998robust}. A general approach to defining the robustness weight function $W_T$ is described below~\cite{doyle2013control}:

\begin{equation}
\left| \frac{M_{ik}e^{j\phi_{ik}}}{M_{i}e^{j\phi_{i}}}-1\right| \leq \left| W_T(j\omega_i)\right|, i=1,\ldots,m;k=1,\ldots,n_r
\label{eqn_wt}
\end{equation}
The magnitude and phase values are assessed over a range of frequencies, denoted as $\omega_i$ (ranging from $i = 1$ to $m$), and the experiment is repeated $n_{r}$ times. The notation $(M_{ik}, \phi_{ik})$ refers to the magnitude-phase measurements corresponding to frequency $\omega_i$ and the $k$th experiment iteration, where $k = 1$ to $n_r$. Similarly, $(M_i, \phi_i)$ represents the magnitude-phase pairs for the nominal plant $P(s)$.

The following remark describes an important dependency of the uncertainty in SPAs:
\begin{remark}\label{rk_2}
The single input (underactuated control) may lead to different motions of soft fingers with the same dimensional parameters due to the uncertainty of the soft materials. The deformation curves of some soft materials (Smooth-on Ecoflex series) exhibit high uncertainty when they have a slow deformation rate~\cite{rebecca2015material, luc2021materials}. In contrast, the curves demonstrate much less uncertainty when their deformation rate is high. Similarly, when a higher pressure changing rate is applied to soft fingers, they show a narrower uncertainty band, align more closely with nominal behaviors, and tend to have consistent motions. This property influences the feasibility and performance of underactuated control of the multi-finger soft gripper (\ref{eqn_simo3}).
\end{remark}

Using the insight from the above remark, equation \eqref{eqn_pertub} can be redefined with respect to the operational speed $\omega_m$. For simplicity, we consider only two fingers:

\begin{footnotesize}
\begin{align*}
&\Pi_H(\omega_m^H) 
:= \Biggl\{ \Biggl( \begin{bmatrix} 1 & 0 \\ 0 & 1 \end{bmatrix} + \Delta \overbrace{\begin{bmatrix} W_{T1} & 0 \\ 0 & W_{T2} \end{bmatrix}}^{W_T^H} \Biggl) \begin{bmatrix} P_1(s) \\ P_2(s) \end{bmatrix} \mid \|\Delta\|_{\infty} \leq \gamma \Biggl\} \\
\\
&\Pi_L(\omega_m^L)
:= \Biggl\{ \Biggl( \begin{bmatrix} 1 & 0 \\ 0 & 1 \end{bmatrix} + \Delta \underbrace{\begin{bmatrix} W_{T1} & 0 \\ 0 & W_{T2} \end{bmatrix}}_{W_T^L} \Biggl) \begin{bmatrix} P_1(s) \\ P_2(s) \end{bmatrix} \mid \|\Delta\|_{\infty} \leq \gamma \Biggl\}
\end{align*}
\end{footnotesize}
such that
\begin{align*}
\omega_m^H > \omega_m^L \implies \bar{\sigma}(W_T^H(j\omega_i)) < \bar{\sigma}(W_T^L(j\omega_i)).
\end{align*}
where $\Pi_H$ denotes the uncertain plant family with respect to high speed of actuation ($\omega_m^H$) and $\Pi_L$  defines the uncertain plants for low speed of actuation ($\omega_m^L$). Thus, as the operational speed of the soft actuator increases, the variance in its behavior is reduced.

\begin{figure}[t]
    \centering
    \includegraphics[width=220pt]{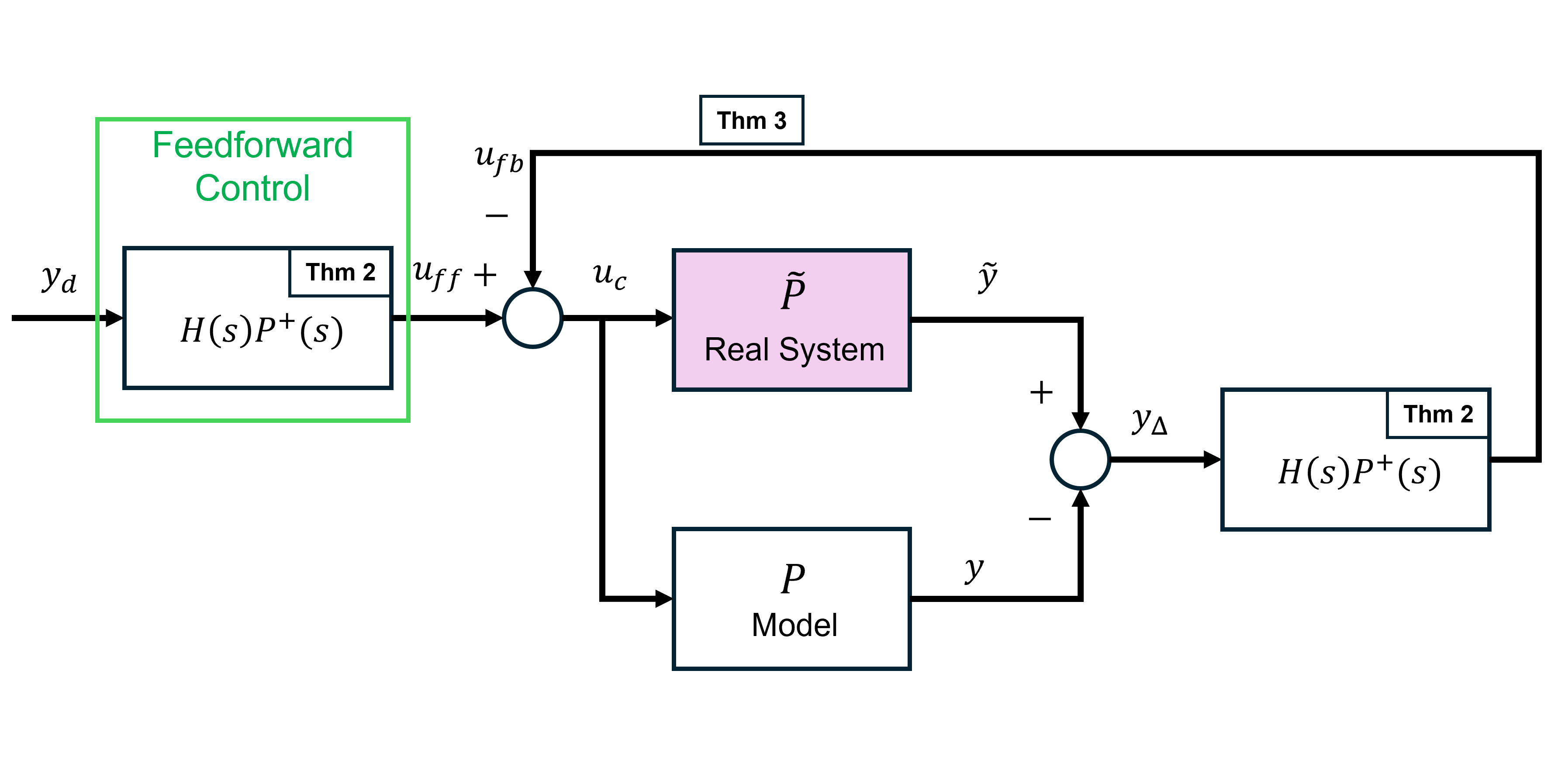}
    \caption{The block diagram of the proposed controller including the feedforward control and feedback loop.}
    \label{fig_6}
\end{figure}

\subsection{Controller Design and Analysis}\label{subsec33}
If all the elements in $P(s)$ of (\ref{eqn_simo2}) are the same (i.e., $P_1 = P_2 = ... = P_{n_y}$), the multiple systems will be coordinated automatically. However, the full system matrix of multiple soft fingers $P(s)$, which has different elements (i.e., $P_1 \neq P_2 \neq ... \neq P_{n_y}$), will be utilized to design the underactuated controller. We can re-formulate (\ref{eqn_simo2}) for the desired output as 

\begin{align}
\begin{bmatrix} Y_{d1}(s)\\ \vdots \\ Y_{d{n_{y}}}(s) \end{bmatrix} = \begin{bmatrix} P_1(s)\\ \vdots \\ P_{n_{y}}(s) \end{bmatrix} {U(s)} 
\label{eqn_simo3}
\end{align}
where $Y_d(s) \in \textnormal{Im}_{\Bbb{R}(s)}(P)$. In (\ref{eqn_simo3}), $Y_{di}(s)$ represents the desired output of $P_{i}(s)$, and $i = 1,...,n_{y}$. The stable inversion is composed of the feedforward and feedback loop~\cite{burak2024algebraic}. The feedforward controller is obtained by solving (\ref{eqn_simo3}) to get $U(s)$ and the additional feedback loop aims to address the system perturbation of the mechatronic system as shown in Fig.~\ref{fig_6}. 

Some results are introduced and will be implemented to design the controller, including the feedforward control and feedback loop.
\begin{theorem}(section III-B, ~\cite{burak2024algebraic})
Let $P(s)$ be non-square ($\textnormal{rank}_{\Bbb{R}(s)}(P) = n_u < n_y$), then there exists an $P\textsuperscript{\textdagger}(s):=(P^T(s)P(s))^{-1}P^T(s)$ satisfying $P\textsuperscript{\textdagger}(s)P(s)=I$. Besides, it is defined that $y_{d}(t)$ is the desired system response in the time domain and $y_{d}^a(t)$ is the system response by applying an approximate solution $U^a(s)$. Thus, an approximate solution $U^a(s)$ is defined as
\begin{equation}
U^a(s) = H(s)P\textsuperscript{\textdagger}(s)Y_{d}(s)
\label{eqn_ff}
\end{equation}
satisfying \\
\indent 1) $H(s) \in \Bbb{R}(s)$\\
\indent 2) $\parallel y_{d}^a(t)-y_{d}(t)\parallel_{\infty}<~\infty~for~t\in [0, \tau]$\\
\indent 3) $y_{d}^a(t) \approx y_{d}(t)~for~t \in (\tau, \infty)$ \\
\indent 4) $U^a(s) \in \mathcal{RH}_{\infty}$
\label{thm_ff}
\end{theorem}

\begin{definition}(section III-B,~\cite{burak2018disturb})
Let $\omega_{cl}$ denote the bandwidth (BW) of the system. So the $H(s)$ is defined as \\
\indent\indent $ H(j\omega) \approx I \Leftrightarrow~\omega \ll \omega_{cl} $\\
\indent\indent $ H(j\omega) \approx 0 \Leftrightarrow~\omega \gg \omega_{cl} $\\
\indent\indent $ H(j\omega) \not\approx \{0, I\} \Leftrightarrow~\omega$ close to $\omega_{cl} $    
\label{filter}
\end{definition}

\begin{remark}
The P\textsuperscript{\textdagger}(s) in (\ref{eqn_ff}) contains non-causal elements. To make it applicable to real systems, the $H(s)$ could be loop-shaping synthesis~\cite{zhou1998robust} or a low-pass filter~\cite{control2002franklin} to re-shape the P\textsuperscript{\textdagger}(s). Considering a general case, a system would be minimal or non-minimal. Loop-shaping synthesis is applied to shape the (\ref{eqn_ff}). However, if the system is minimal as stated in the Assumption~\ref{ass_minimumsystem} and has a lower order, the low-pass filter, which is relatively more applicable, can be utilized to shape the (\ref{eqn_ff}). The order of the low-pass filter depends on the relative order ($l$) of P\textsuperscript{\textdagger}(s). The low-pass filter takes the form of $\frac{a_0}{s^l + a_{l-1} s^{l-1} + ... + a_0}$. The parameters $a_0, \hdots a_{l-1}$ in the equation are selected to define the cut-off frequency of the low-pass filter. 
\end{remark}
\begin{remark}
An alternative reason to apply $H(s)$ is that using the P\textsuperscript{\textdagger}(s) in (\ref{eqn_ff}) directly in the feedforward or feedback controllers can cause undesired high-frequency excitation. Thus, an appropriate selection of the either cut-off frequency of the low-pass filter or the bandwidth of the resulting complementary sensitivity function of the loop shaping can prevent this excitation. As demonstrated in the experimental part, there is a definite advantage of letting the bandwidth of $H(s)$ be as large as possible so that the better synchronization of multiple fingers be achieved due to the nonlinear nature of the material of soft fingers.
\end{remark}

The tracking error (with perturbed term) can be compensated by using the output feedback as displayed in Fig.~\ref{fig_6}. The $u_{ff}$ is calculated based on the \textit{Theorem~\ref{thm_ff}}. It is assumed that the output of the real (uncertain) system can be measured such as sensors or observers. Since the nominal system output can be computed with the combined input $u_c$, we have the output difference $y_{\Delta}(t)$. With this output difference, we can compensate for the error by the following theorem.
\begin{theorem}~\cite{burak2024algebraic}
Consider the block diagram in Fig.~\ref{fig_6} with (\ref{eqn_pertub}). Then the bounded $U_{fb}$ yields $\parallel y_d(t) - \tilde{y}(t)\parallel_{\infty}\rightarrow 0$ $\textit{iff}~\tilde{Y}(s) \in \textnormal{Im}_{\Bbb{R}(s)}(P)$, where $y_d(t)$ has the same definition in \textit{Theorem~\ref{thm_ff}} and $\tilde{y}$ is the real system response with feedback loop.
\label{thm_fb}
\end{theorem}

For notational simplicity, we take ${\Delta}{W_T}$ in (\ref{eqn_pertub}) as $\Delta_P$, and consider $\tilde{P}(s) \in \Pi$, $Y_d = PU_{ff}$, and $U_c(s):=U_{ff}-U_{fb}$ (by referencing Fig.~\ref{fig_6})
\begin{equation}
\tilde{P}(s)U_c(s) = \tilde{Y}(s)=PU_c+\Delta_P PU_c
\label{eqn_thm91}
\end{equation}
\begin{align}
&PU_{ff}+\Delta_P PU_{ff}-PU_{fb}-\Delta_P PU_{fb}=\tilde{Y}(s)\\
&\Rightarrow PU_{fb}=\Delta_P PU_{ff}-\Delta_P PU_{fb}
\label{eqn_thm92}
\end{align}
The feedback loop and $U_{fb}$ will compensate for the model errors ($\Delta_P$). The proof of \textit{Theorem \ref{thm_ff}} and \textit{\ref{thm_fb}} are given in~\cite{burak2024algebraic}.

The model inversion and feedback loop are introduced to achieve accurate tracking of the system. The next problem is whether the system performance can be achieved by underactuated control as depicted in (\ref{eqn_simo3}). That is, multiple soft actuators are controlled with a single input pressure. Here, the goal is to control the motions of the multi-finger gripper to reach stable grasping~\cite{yang2023control}. The desired response of each finger is assigned as $Y_{d}$ in (\ref{eqn_simo3}) and the ${Y}_d(s) \in \textnormal{Im}_{\Bbb{R}(s)}(P)$. This problem will be shown to be solvable and the solution to exist through mathematical inference. Since the $P(s)$ has full rank = 1, the $P(s)$ is invertible and exists left inverse matrix according to the \textit{Theorem~\ref{thm_ff}}. The pseudo-inverse is $P\textsuperscript{\textdagger}(s) = (P^{T}(s)P(s))^{-1}P^{T}(s)$. The controller is obtained by $U^a(s) = H(s)P\textsuperscript{\textdagger}(s){Y}_d(s)$. Therefore, there exists input $U(s)$ that makes the ${Y}_d(s)$ achievable.

\begin{figure}[t]
    \centering
    \includegraphics[width=220pt]{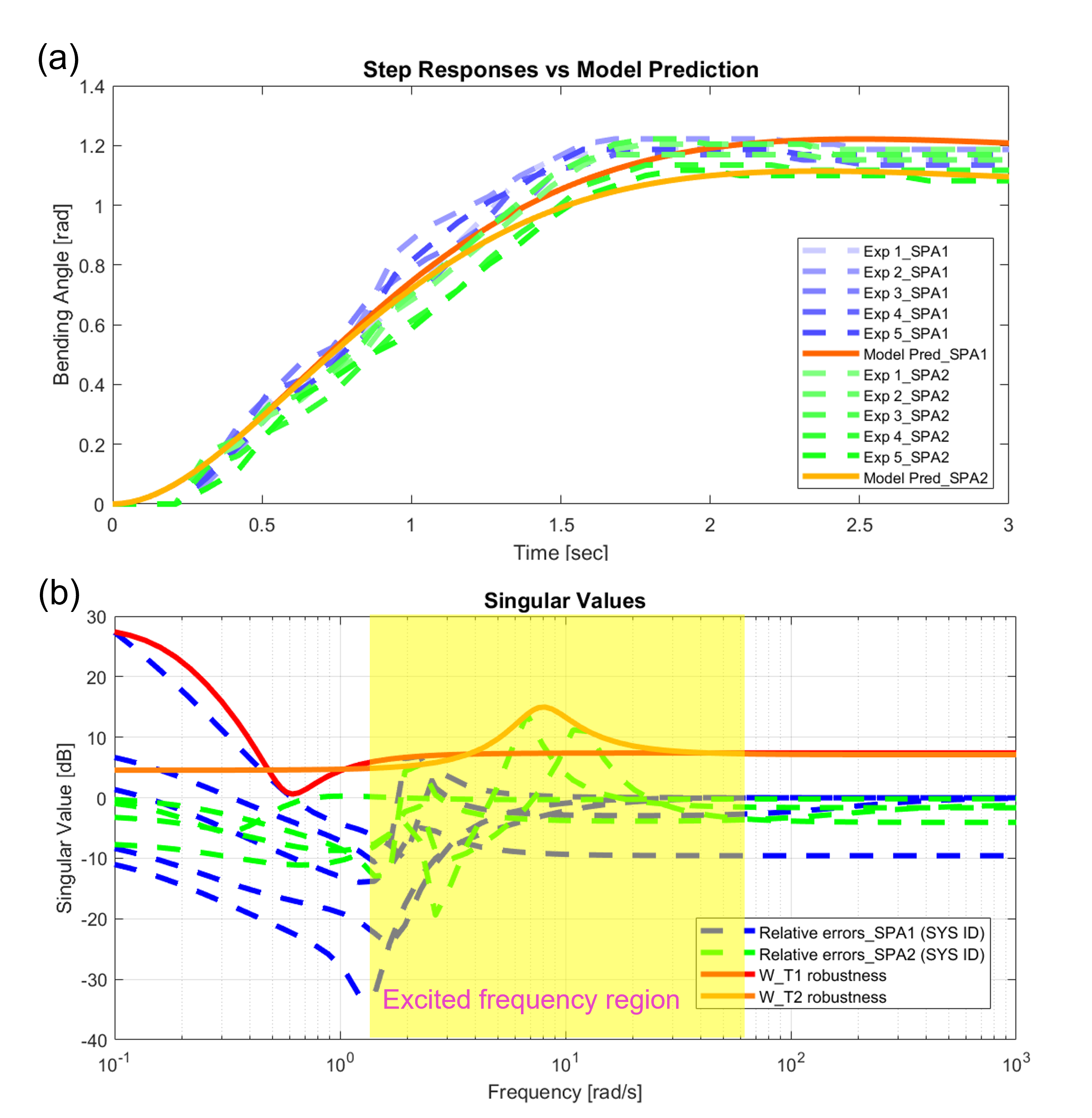}
    \caption{Several open-loop responses of both soft pneumatic actuators (soft fingers) are demonstrated in (a). The robustness weight selection of both soft fingers based on the modeling errors can be seen in (b). }
    \label{fig_9}
\end{figure}

\section{Experimental Evaluation}\label{sec4}
In previous sections, we introduce the mechatronic system, specifically the multi-finger soft gripper, alongside system dynamical models and control algorithms. Prior to experimentation, preliminary tests are conducted using MATLAB\textregistered/Simulink to assess the feasibility of the stable inversion algorithm. Subsequently, a series of experiments are executed to evaluate the practicality of the proposed control approach. An additional disturbance test is then performed to evaluate the robustness of the controller.

\subsection{Preliminary Evaluations}\label{pre_eval}
Some definitions and assumptions are applied in Sec.~\ref{sec2} and~\ref{sec3}. This subsection intends to evaluate that the definitions and assumptions are valid before the simulations and experimentation.
\subsubsection{Model Evaluation}
The analytical model matrix is built for the mechatronic system as $P(s)$ in~(\ref{eqn_simo5}). Based on the discussion in Sec.~\ref{subsec33}, the model has uncertainty and will cause modeling errors due to the uncertain soft materials. The bounded model uncertainty is assumed, $\parallel \Delta \parallel_{\infty}\leq 1 $ in (\ref{eqn_pertub}). Several step responses of both soft fingers are conducted to evaluate whether the model uncertainty is bounded.

\begin{figure}[t]
    \centering
    \includegraphics[width=220pt]{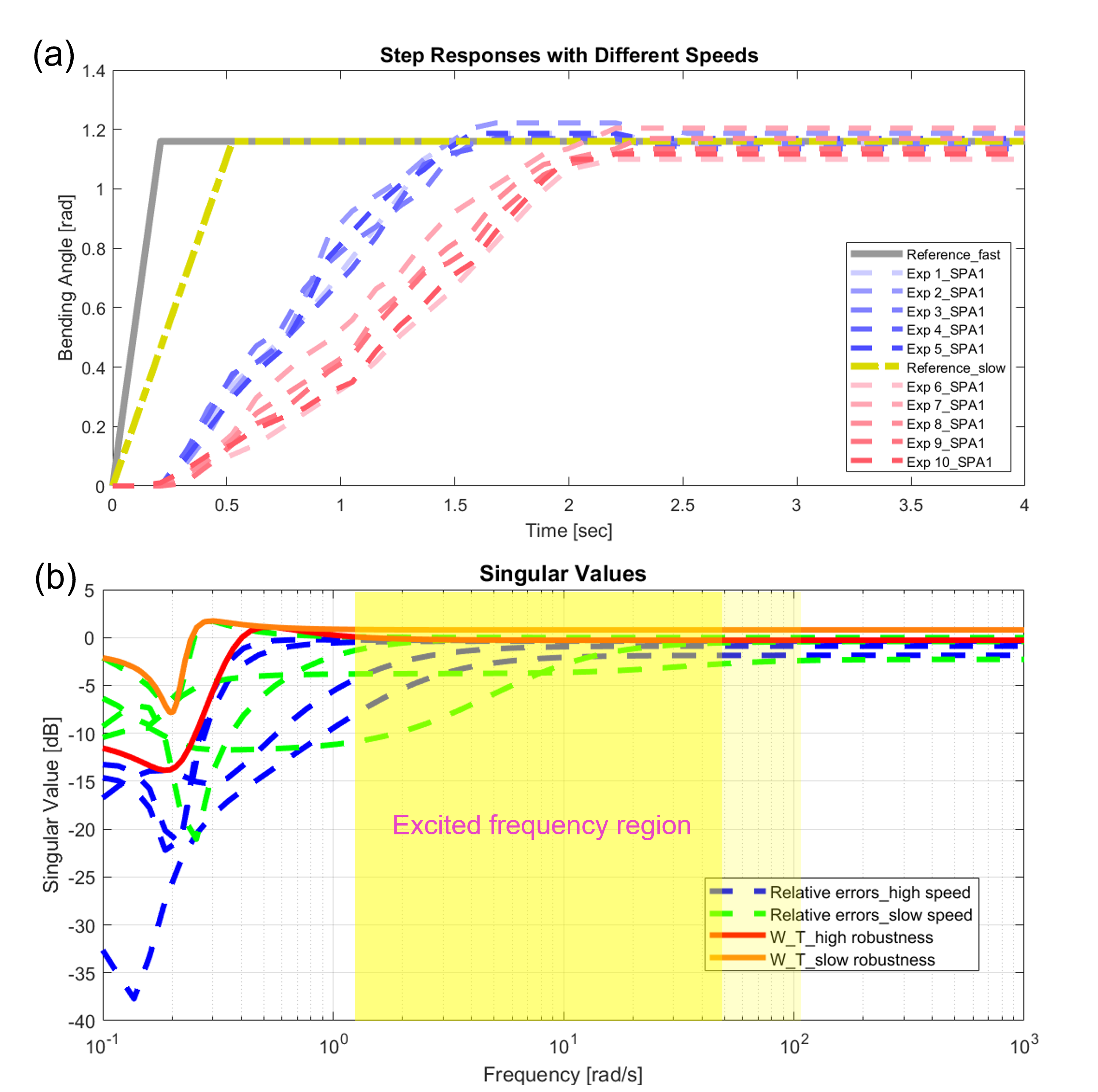}
    \caption{The step responses of different speeds of SPA1 (left finger in Fig.~\ref{fig_1}) are shown in (a). The robustness weight selection of SPA1 with high and slow speeds based on the modeling errors can be seen in (b).}
    \label{fig_8}
\end{figure}

Figure~\ref{fig_9} (a) demonstrates the repeated step responses of the two soft actuators, the first (blue) and second (green) element of the $P(s)$ in~(\ref{eqn_simo5}). The system performance varies due to the properties of soft materials~\cite{yang2023control}. The singular values of the bounded constraint in (\ref{eqn_pertub}) can be found in Fig.~\ref{fig_9} (b). The system perturbation of two soft actuators is bounded. Specifically, based on the (\ref{eqn_uncertain}), it is observed that $\parallel C_n - C_{nominal} \parallel~\leq 14.3 \%$ and $\parallel K_n - K_{nominal} \parallel~\leq 5.9 \%$ by performing system identification of the step responses in Fig.~\ref{fig_9} (a). The $C_{nominal}$ and $K_{nominal}$ are the average of the identified model of those multiple step responses. Two soft actuators, both elements of the $P(s)$ in~(\ref{eqn_simo5}), show a similar result. It is concluded that the perturbations of the two systems are bounded, confirming the validity of the model evaluation. This evaluation holds within the excited frequency range from 6 to 63 $rad/s$ as the yellow region in Fig.~\ref{fig_9} (b), determined by performing the Fast Fourier transform on the input reference as shown in Fig.~\ref{fig_9} (a).
\subsubsection{Controllability Evaluations}
Another preliminary evaluation is needed before the experimentation. The system equation of (\ref{eqn_simo2}) should be controllable and observable to ensure that the system realization is minimal. Note that if the two fingers are identical, the realization is uncontrollable. In this case, the motion of the two fingers is always synchronized. If the controllability matrix $M_c = [A~BA~BA^2~\cdots]$ and the observability matrix $M_o = [C~CA~CA^2~\hdots]^T$ of (\ref{eqn_simo2}) both have full rank, the full system is controllable and observable. Since the full system is controllable and observable, the system realization is also minimal. This evaluation matches the \textit{Assumption~\ref{ass_minimumsystem}. a}. The proposed controller design is valid.

\begin{figure}[t]
    \centering
    \includegraphics[width=200pt]{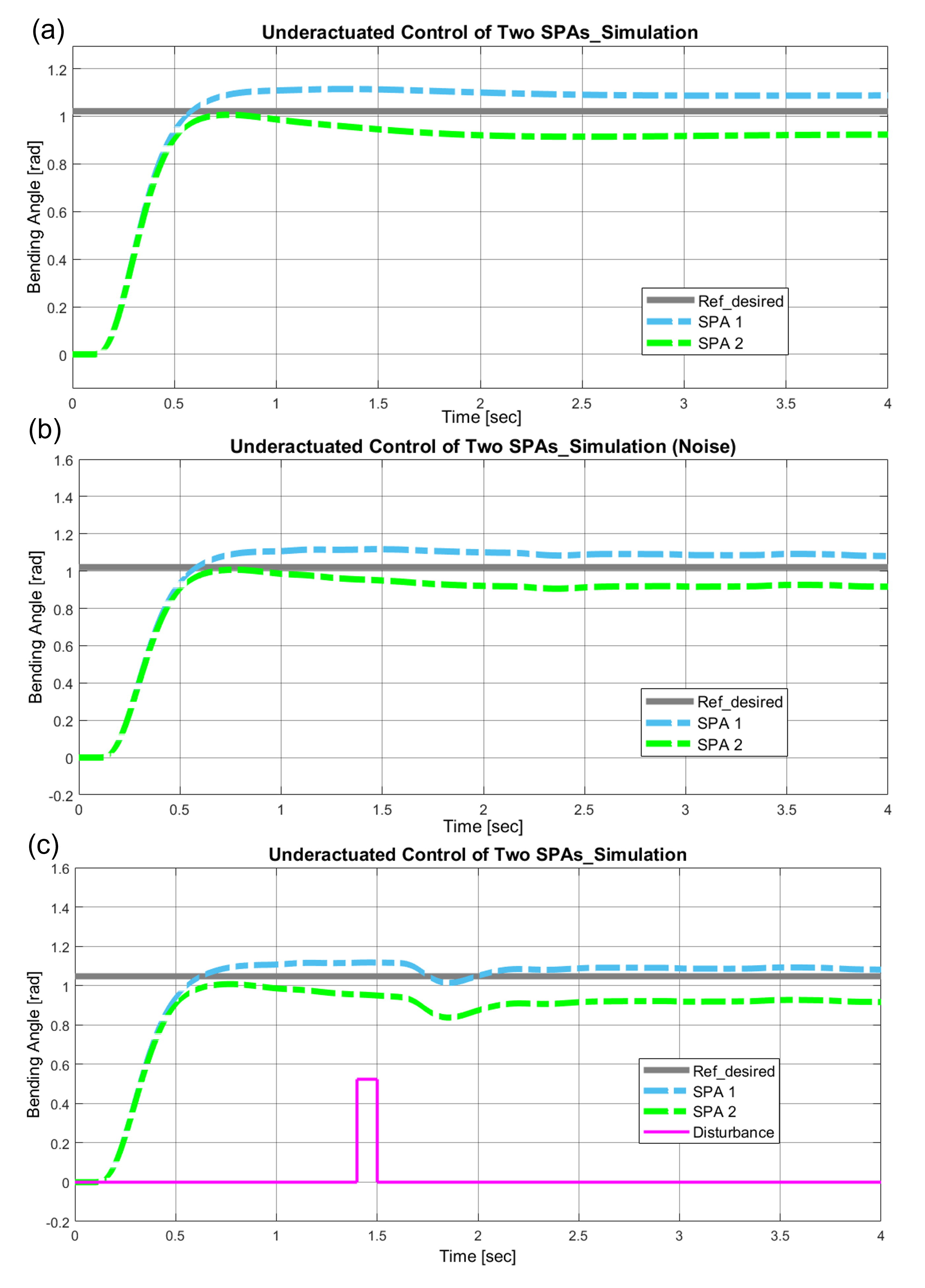}
    \caption{The simulation results of the under-actuated control of the two-finger gripper are depicted in (a). The sensor noise and external disturbance are considered and visualized in (b) and (c).}
    \label{fig_7}
\end{figure}

\subsubsection{Uncertainty Band of Soft Finger}\label{uncertainband}
It is observed that the soft materials exhibit larger uncertainty as discussed in \textit{Remark~\ref{rk_2}} of Sec.~\ref{subsec33}. The experimental evaluation of this property is conducted and visualized in Fig.~\ref{fig_8} (a) and (b). The left finger in Fig.~\ref{fig_1} is used to conduct this evaluation. Two input commands with different speeds are applied to do step response tests of a soft finger. The slower-speed input command excites a larger uncertainty band (red lines). The standard deviation of the steady-state error is 2.41 $deg$, while that of higher-speed input command is 1.14 $deg$ (blue lines). The input references in Fig.~\ref{fig_8} (a) are converted to the frequency domain using the Fast Fourier transform to validate the excited frequency region, as the yellow region in Fig.~\ref{fig_8} (b). This soft actuator's frequency region spans from approximately 6 to 50 $rad/s$. Figure~\ref{fig_8} (b) illustrates that the magnitude variations of the soft actuator converge at higher frequencies, approaching the nominal dynamics. These results confirm that the soft actuator demonstrates reduced uncertainty and approximates nominal dynamics at higher speeds (or frequencies), consistent with the observations in the time domain. This property plays a vital role in the multiple soft finger underactuated control since soft fingers have different dynamical models. Higher speeds may reduce the uncertainty band and help coordinate multiple soft fingers within a soft gripper. 

\subsection{Simulation Results}\label{subsec41}
Section \ref{subsec22} introduces the general case of the problem (\ref{eqn_simo2}). The system matrix $P(s)$ comprises $n_{y}$ elements. Simulations will be conducted on multi-finger soft gripper systems to assess different scenarios, with $n_{y}$ taking values of 2 for $P(s)$. The simulations help us understand if the controller can work and regulate the system output. Besides, the low pass filter $H(s)$ of \textit{Theorem \ref{thm_ff}} will be adjusted to achieve better tracking performance.



The simulation is performed on a two-finger gripper, which includes two soft actuators (soft fingers) and a single syringe pump. The ideal desired output is selected as a step response ${Y}_d = [\frac{0.333\pi}{s}, \frac{0.333\pi}{s}]^T$. As the elements in $P(s)$ are nonidentical in (\ref{eqn_simo5}), the ${Y}_d$ is not in the image space of $P$ and it can be factorized as~\cite{burak2024algebraic} 

\begin{align}
Y_d = \underset{\text{Im}(P)}{\textnormal{proj}}(Y_d) + \textnormal{res}(Y_d) 
\label{eqn_res}
\end{align}
\begin{align}
\underset{\text{Im}(P)}{\textnormal{proj}}(Y_d) = \sum_{n=1}^{r} \langle {Y_d}, {q_i} \rangle  {q_i}
\label{eqn_gs}
\end{align}
where ${q_i}s$ is defined in (\ref{eqn_image}). The $\textnormal{res}(Y_d)$ can be obtained by $Y_d - \textnormal{proj}(Y_d)$ based on (\ref{eqn_res}) and (\ref{eqn_gs}). Note that this obtained desired output vector may lead to the collision of the two soft fingers as Fig.~\ref{fig_1}, but we aim to check the performance of the control algorithm. With the $\textnormal{res}(Y_d)$, $\textnormal{proj}(Y_d) = [0.303\pi/s, 0.357\pi/s]^T$ is in the range space of $P$. The system equation is written as

\begin{align}
\begin{bmatrix} \frac{0.303 \pi}{s}\\ \frac{0.357 \pi}{s} \end{bmatrix} &= \begin{bmatrix} \frac{7.831}{s^3 + 2.66 s^2 + 3.61 s} \\ \frac{7.831}{s^3 + 2.45 s^2 + 3.06 s} \end{bmatrix} {U(s)} 
\label{eqn_simo5}
\end{align}
The systems in (\ref{eqn_simo5}) will have steady-state tracking errors, $0.030\pi$ (SPA 1) and $-0.024\pi$ $rad$ (SPA 2) respectively compared to the desired reference ($\pi/3~rad$). The controller is obtained based on the \textit{Theorem~\ref{thm_ff}} and \textit{\ref{thm_fb}}, so the $P\textsuperscript{\textdagger}(s) = [\frac{s^3 + 2.66 s^2 + 3.61 s}{15.66},~\frac{s^3 + 2.45 s^2 + 3.06 s}{15.66}]$ and the $H(s)$ is designed by loop-shaping the element of P\textsuperscript{\textdagger}(s) with desired equation $4.8/s$, and 
\begin{small}
$H(s) = \frac{1009s^3 + 2.52e5s^2 + 6.829e5s + 1.027e6}{s^6 + 282.7s^5 + 6.14e3s^4 + 8.022e4s^3 + 4.335e5s^2 + 8.948e5s + 1.027e6}. $
\end{small}

This controller is able to coordinate the motions of the two fingers within the soft gripper as Fig.~\ref{fig_7} (a). The settling time of each finger is approximately 0.6 $sec$, making the soft gripper comparable to traditional grippers. The tracking error of the two fingers is within 1 $deg$ compared to the $\textnormal{proj}(Y_d)$. To further evaluate the performance of the controller, sensor noise and disturbance are added as Fig.~\ref{fig_7} (b) and (c). The feedback loop can compensate for the sensor noise as Fig.~\ref{fig_7} (b). The root-mean-square error (RMSE) of the sensor noise is set as $\pm$ 2 $deg$~\cite{yang2023control}. The RMSE of the two fingers is within 0.5 $deg$, so the sensor noise does not affect the systems' output.

The disturbance here is regarded as the soft fingers are hitting by an external force. The results show that the controller is capable of adjusting the system back to the reference as Fig.~\ref{fig_7} (c). The amplitude of the errors for both fingers is within 0.1 $rad$, so the feedback control can handle the disturbance. The simulation results validate the \textit{Theorem \ref{thm_fb}} and (\ref{eqn_thm92}). The experimental results can be referenced in Sec.~\ref{subsec44} and \ref{subsec45}.

The simulation results on two-finger soft grippers endorse the feasibility of the proposed controller algorithm. The motions of fingers are coordinated. These dynamics will be beneficial to grasping tasks when the soft gripper is applied to manipulate various objects. The controller will be applied to the real soft gripper to evaluate the applicability of this control algorithm.

\subsection{Experimental Setup}\label{subsec42}
Figure \ref{fig_1} illustrates both the experimental arrangement and the signal flow diagram. The two-finger soft gripper is used to conduct the experiments with a single syringe pump. The soft actuators are fabricated by molds as illustrated in Fig.~\ref{fig_2} (b). The motions of the soft fingers are driven by the syringe pump as Fig.~\ref{fig_1}~\cite{yang2023pump}, which is actuated by a stepper motor. A DM320T digital stepper driver (StepperOnline, New York, NY) is utilized to trigger the stepper motor. An air pressure sensor (Walfront, Lewes, DE) with a sensing range of 0 to 80~$psi$ is utilized to detect the air pressure for open-loop control. Additionally, each soft finger contains a flex sensor (Walfront, Lewes, DE) inside to monitor its bending angle, facilitating feedback control. The flex sensor is a resistive type sensor and has a sensing range of 100 $deg$ and a sensing error of approximately 2 $deg$ (root-mean-square error). Both sensors are synchronized with Arduino MEGA 2560 (SparkFun Electronics, Niwot, CO), which is based on the Microchip ATmega 2560. The controller algorithms are programmed in MATLAB\textregistered/Simulink which is communicated with the Arduino MEGA 2560 to process feedback signals and generate control commands for the mechatronic system. The model and controller are discretized in the analytical software with a sampling time of 100 $ms$. The \enquote{Real System} block in Fig.~\ref{fig_6} is replaced by the real soft gripper.

\begin{figure}[t]
    \centering
    \includegraphics[width=200pt]{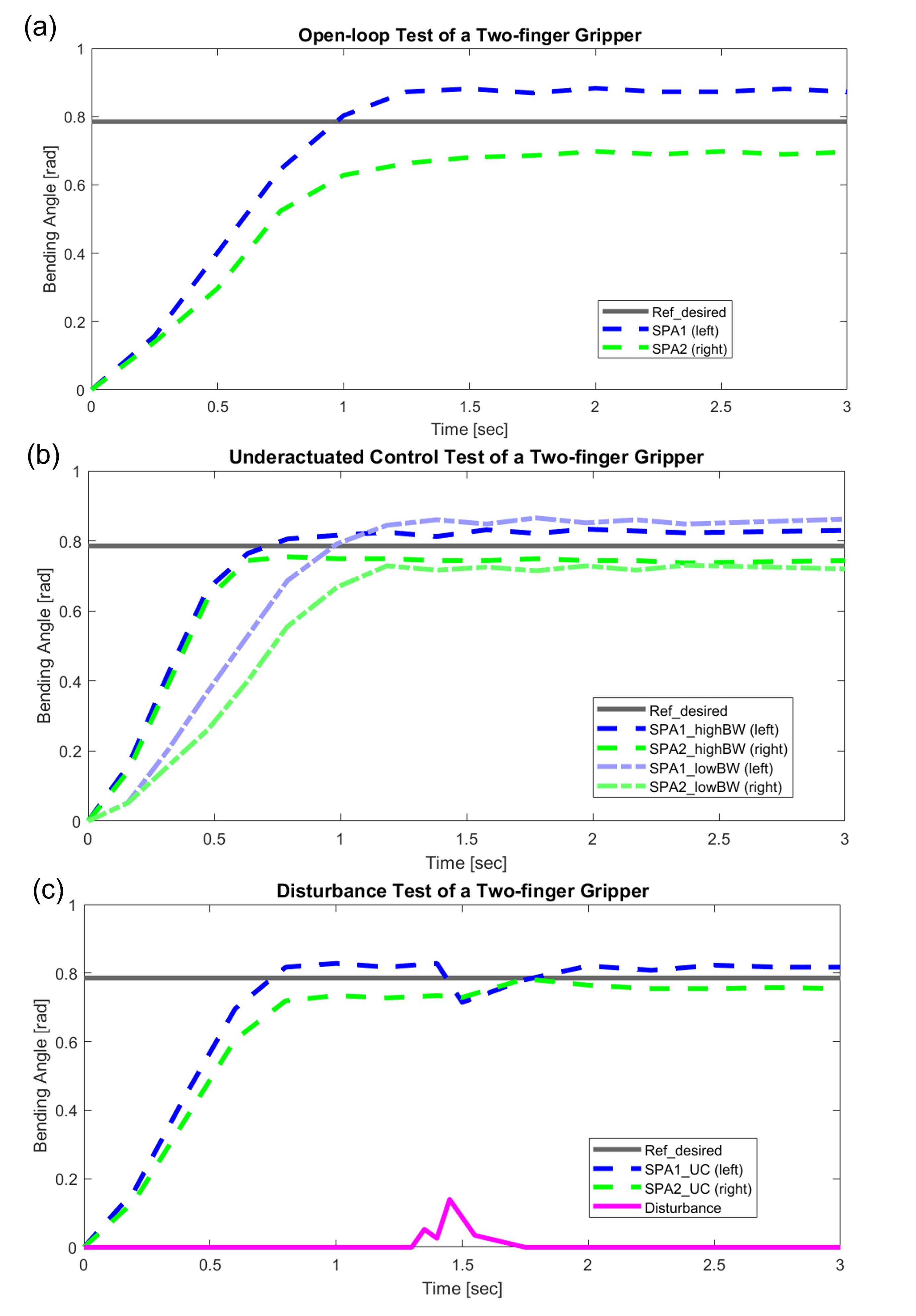}
    \caption{The visualization of the open-loop test of the two-finger gripper is displayed in (a). The responses of using the proposed controllers designed for different bandwidths are demonstrated in (b). The motions of the two fingers are coordinated compared to the results in (a). The disturbance test is depicted in (c) and the controller can handle the external disturbance.}
    \label{fig_10}
\end{figure}

\subsection{Open-loop \& Closed-loop Tests}\label{subsec43}
Prior to implementing the proposed controller, the open-loop test attempts to visualize the open-loop responses of the soft fingers. 
The soft fingers share the same dimensional parameters such as height, width, and length. However, their system parameters in equation (\ref{eqn_22}) differ, resulting in distinct motions. In applications, the open-loop control results in inconsistent motions of multiple fingers which is not beneficial for manipulation tasks. The open-loop test results are illustrated in Fig.~\ref{fig_10} (a). 

The desired reference is set as 45 $deg$ ($\pi$/4 $rad$) to avoid real collision between two fingers. Based on the (\ref{eqn_res}) and (\ref{eqn_gs}), the desired references for left and right fingers are 48.24 and 40.86 $deg$, respectively. The blue dashed line represents the response of the left finger (SPA 1) in Fig.~\ref{fig_1} while the green dashed line denotes the right finger (SPA 2) in Fig.~\ref{fig_1}. The left finger is active and responds faster. By contrast, the right finger has relatively slow responses. Their steady states are also different due to different system parameters of~(\ref{eqn_simo5}). The left finger reaches approximately 50 $deg$ while the right finger reaches around 39.5 $deg$.

\subsection{Underactuated Control Tests}\label{subsec44}
The stable inversion algorithm in Sec.~\ref{subsec33} is implemented to control the two fingers in this subsection. The system equation is similar to (\ref{eqn_simo5}) in Sec.~\ref{subsec41} and is described as 

\begin{align}
\begin{bmatrix} \frac{0.227 \pi}{s}\\ \frac{0.268 \pi}{s} \end{bmatrix} &= \begin{bmatrix} \frac{7.831}{s^3 + 2.66 s^2 + 3.61 s} \\ \frac{7.831}{s^3 + 2.45 s^2 + 3.06 s} \end{bmatrix} {U(s)} 
\label{eqn_simo6}
\end{align}
where $P\textsuperscript{\textdagger}(s)$ and $H(s)$ are the same as shown in Sec.~\ref{subsec42}, and the desired reference is set as 45 $deg$ ($\pi$/4 $rad$). According to the (\ref{eqn_res}) and (\ref{eqn_gs}), the $Y_d = [0.25\pi/s, 0.25\pi/s]^T$ is factorized as $\textnormal{proj}(Y_d)$ and $\textnormal{res}(Y_d)$. The $\textnormal{proj}(Y_d) = [0.227\pi/s, 0.268\pi/s]^T$ is in the range space of $P(s)$, and the steady-state tracking errors of systems in (\ref{eqn_simo6}) are $0.023\pi$ and $-0.018\pi$ $rad$ respectively compared to the desired reference $\pi/4~rad$.

The result is demonstrated in Fig.~\ref{fig_10} (b). Two soft fingers reach their steady states at nearly the same time, and their tracking errors are below 1 $deg$ compared to the $\textnormal{proj}(Y_d)$. The settling time is around 0.7 $sec$, which is better than our previous research by using an optimal controller~\cite{yang2023control}. The response time enables the soft gripper to be comparable to rigid grippers. Besides, their transient states are nearly synchronized with around 2 $deg$ differences, which support the evaluation of Sec.~\ref{uncertainband}. 

Additionally, the bandwidth of $H(s)$ affects the systems' responses, so another controller is designed whose $H(s)$ has narrower bandwidth. The $H(s)$ is designed by loop-shaping the element of P\textsuperscript{\textdagger}(s) with desired equation $2/s$, and 
\begin{small}
$H(s) = \frac{49.47s^3 + 1991s^2 + 4917s + 6270}{s^6 + 51.21s^5 + 442.5s^4 + 2339s^3 + 6400s^2 + 9281s + 6270}. $
\end{small}
The experimental results are also shown in Fig.~\ref{fig_10} (b). The response of each finger is slower, and the settling time is around 1.1 $sec$. The error of each finger is larger compared to the performance of the controller with a larger bandwidth.

Due to the proposed controller, system response is quicker and reduces the uncertainty band of soft actuators. Our previous work~\cite{yang2023control} utilized a syringe pump for each soft finger of the two-finger gripper to reach synchronization. However, one more syringe pump is needed compared to this research. If there are three-finger or four-finger grippers, more syringe pumps are required, which implies more costs, space, and weight. That makes the applications of the soft gripper setup more difficult.

\subsection{Disturbance Tests}\label{subsec45}
According to the \textit{Theorem~\ref{thm_fb}}, the feedback loop is designed to deal with the model errors or disturbances caused by external forces based on the \textit{Theorem~\ref{thm_fb}} ((\ref{eqn_thm91}) and (\ref{eqn_thm92})). The disturbance is generated by a human finger and is given at around t = 1.4 $sec$ when the fingers arrive at the steady state as Fig.~\ref{fig_10} (c). The external force is only applied to the left finger. The proposed control algorithm is able to regulate the systems to the desired reference when the external force is applied. The experimental results support the \textit{Theorem~\ref{thm_fb}}. 

\section{Discussion and Conclusion}\label{sec5}
\subsection{Discussion}\label{subsec1}

The proposed control algorithms successfully coordinate the motions of two soft fingers within a soft gripper. The performance is validated by real-world experimentation. However, if systems have identical models (i.e., $P_1 = P_2 = ... = P_{n_y}$ in (\ref{eqn_simo2})), those systems will automatically coordinate their motions given an input $U(s)$. Since the models of soft fingers are different (i.e., $P_1 \neq P_2 \neq ... \neq P_{n_y}$ in (\ref{eqn_simo2})), this approach is proposed to address this issue and coordinate their motions. Even if the soft fingers have the same dimensional parameters, their system models are different due to the properties of soft materials~\cite{rebecca2015material}.

Furthermore, there is a limitation due to the hardware configuration of the soft gripper. The soft gripper has a parallel nature of multiple fingers driven by a single syringe pump. If the desired output function is out of the image space of $P(s)$, the soft gripper may not reach the desired states. For instance, if the desired function of one finger is $\pi/4~rad$ and another one is $-\pi/4~rad$, the solution of the system (\ref{eqn_simo3}) does not exist. The negative bending angle is out of the image space of this soft gripper.

The proposed approach sets the input-output relation characterized by $H(s)$, either a low-pass filter or designed by loop-shaping. The system responses are influenced by the $H(s)$. By enlarging the bandwidth of $H(s)$, the output moves toward the steady-state angle faster. Thus, the system responses are optimized by designing a suitable $H(s)$ as shown in both Sec.\ref{subsec41} and Sec.~\ref{subsec44}. The simulation and experimental visualizations are depicted in Fig.~\ref{fig_7} (a) and Fig.~\ref{fig_10} (b). The steady-state angle of each finger comes closer to the desired angle, and as a result, better coordination is achieved.

\subsection{Conclusion}\label{subsec2}
This paper explores the underactuated control of multiple fingers within a soft gripper, validating a controller that contains feedforward and feedback loops designed via stable model inversion. The soft fingers are designed based on an optimal design framework and their dynamical models are obtained by applying nonlinear mechanics. The feedforward controller is designed based on stable inversion of the system model matrix, while the feedback loop is incorporated to handle the system perturbations. Simulation results demonstrate the efficacy of the control algorithms in controlling a two-finger gripper. Experimental validation further confirms the feasibility of coordinating motions within a two-finger gripper setup, achieving high-speed transient responses and minimal steady-state errors. Even if there is a disturbance, the controller is able to regulate the systems to the reference. The control strategy reduces the number of inputs (air pumps) which may benefit the implementation of multi-finger soft grippers.

In the future, the proposed control theories will be extended to broader cases. The desired output functions are now restricted to the range space of the system model. The extended research will discuss if any solution exists when the desired output functions are out of the image space of the system model theoretically and experimentally.

\section*{Acknowledgments}
The authors would like to thank NSK, Ltd. for arranging the linear actuators used in the experiments.

%


\bibliographystyle{IEEEtran}
\bibliography{IEEEabrv}

\vfill

\end{document}